\def\year{2020}\relax
\documentclass[letterpaper]{article} 
\usepackage{aaai20}  
\usepackage{times}  
\usepackage{helvet} 
\usepackage{courier}  
\usepackage[hyphens]{url}  
\usepackage{graphicx} 
\usepackage{graphicx}  
\usepackage{microtype}
\usepackage{booktabs}
\usepackage{tabularx}
\usepackage{amssymb}
\usepackage{amsmath}
\usepackage{amsfonts}
\usepackage{times}
\usepackage{latexsym}
\usepackage{graphicx}
\usepackage{url}
\usepackage{accents}

\newcommand{\citet}[1]{\citeauthor{#1} \shortcite{#1}}
\newcommand{\citep}{\cite}
\newcommand{\newcite}{\citet}

\nocopyright

\title{Two-Level Transformer and Auxiliary Coherence Modeling \\ for Improved Text Segmentation}

\author{\Large \textbf{Goran Glava\v{s}\textsuperscript{\rm 1} and Swapna Somasundaran\textsuperscript{\rm 2}} \\ 
\textsuperscript{\rm 1}Data and Web Science Research Group\\
University of Mannheim \\
goran@informatik.uni-mannheim.de \\
\textsuperscript{\rm 2}Educational Testing Service (ETS) \\
ssomasundaran@ets.org
}

\newcommand\thickbar[1]{\accentset{\rule{.4em}{.8pt}}{#1}}

\begin{document}

\maketitle

\begin{abstract}
Breaking down the structure of long texts into semantically coherent segments makes the texts more readable and supports downstream applications like summarization and retrieval. Starting from an apparent link between text coherence and segmentation, we introduce a novel supervised model for text segmentation with simple but explicit coherence modeling. Our model -- a neural architecture consisting of two hierarchically connected Transformer networks -- is a multi-task learning model that couples the sentence-level segmentation objective with the coherence objective that differentiates correct sequences of sentences from corrupt ones. The proposed model, dubbed Coherence-Aware Text Segmentation (CATS), yields state-of-the-art segmentation performance on a collection of benchmark datasets. Furthermore, by coupling CATS with cross-lingual word embeddings, we demonstrate its effectiveness in zero-shot language transfer: it can successfully segment texts in languages unseen in training.    
\end{abstract}

\section{Introduction}


Natural language texts are, more often than not, a result of a deliberate cognitive effort of an author and as such consist of semantically coherent segments. Text segmentation deals with automatically breaking down the structure of text into such topically contiguous segments, i.e., it aims to identify the points of topic shift \cite{hearst1994multi,choi2000advances,brants2002topic,riedl2012topictiling,du2013topic,glavavs2016unsupervised,koshorek2018text}. Reliable segmentation results with texts that are more readable for humans, but also facilitates downstream tasks like automated text summarization \cite{angheluta2002use,bokaei2016extractive}, passage retrieval \cite{huang2003applying,shtekh2018exploring}, topical classification \cite{zirn2016classifying}, or dialog modeling \cite{manuvinakurike2016toward,zhao2017joint}.

\begin{figure}
    \centering
    \includegraphics[scale=0.7]{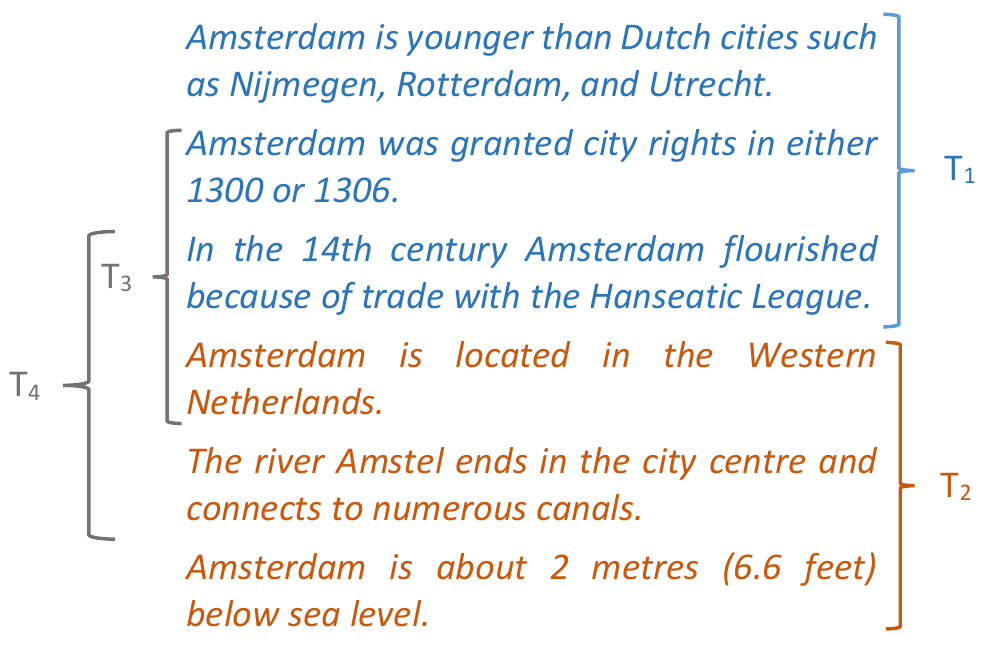}
    \caption{Snippet illustrating the relation (i.e., dependency) between text coherence and segmentation.}
    \label{fig:snippet}
\end{figure}

Text coherence is inherently tied to text segmentation -- intuitively, the text within a segment is expected to be more coherent than the text spanning different segments. Consider, e.g., the text in Figure \ref{fig:snippet}, with two topical segments.    
Snippets $T_1$ and $T_2$ are more coherent than $T_3$ and $T_4$: all $T_1$ sentences relate to \textit{Amsterdam's history}, and all $T_2$ sentences to \textit{Amsterdam's geography}; in contrast, $T_3$ and $T_4$ contain sentences from both topics. $T_1$ and $T_2$ being more coherent than $T_3$ and $T_4$ signals that the fourth sentence starts a new segment.        

Given this duality between text segmentation and coherence, it is surprising that the methods for text segmentation capture coherence only implicitly. Unsupervised segmentation models rely either on probabilistic topic modeling \cite{brants2002topic,riedl2012topictiling,du2013topic} or semantic similarity between sentences \cite{glavavs2016unsupervised}, both of which only indirectly relate to text coherence. Similarly, a recently proposed state-of-the-art supervised neural segmentation model \cite{koshorek2018text} directly learns to predict binary sentence-level segmentation decisions and has no explicit mechanism for modeling coherence.

In this work, in contrast, we propose a supervised neural model for text segmentation that explicitly takes coherence into account: we augment the segmentation prediction objective with an auxiliary coherence modeling objective. Our proposed model, dubbed Coherence-Aware Text Segmentation (CATS), encodes a sentence sequence using two hierarchically connected Transformer networks \cite{vaswani2017attention,devlin2018bert}.
Similar to \cite{koshorek2018text}, CATS' main learning objective is a binary sentence-level segmentation prediction. However, CATS augments the segmentation objective with an auxiliary coherence-based objective which pushes the model to predict higher coherence for original text snippets than for corrupt (i.e., fake) sentence sequences. We empirically show (1) that even without the auxiliary coherence objective, the Two-Level Transformer model for Text Segmentation (TLT-TS) yields state-of-the-art performance across multiple benchmarks, (2) that the full CATS model, with the auxiliary coherence modeling, further significantly improves the segmentation, and (3) that both TLT-TS and CATS are robust in domain transfer. Furthermore, we demonstrate models' effectiveness in zero-shot language transfer. Coupled with a cross-lingual word embedding space,\footnote{See \cite{Ruder2018survey,glavas2019properly} for a comprehensive overview of methods for inducing cross-lingual word embeddings.} our models trained on English Wikipedia successfully segment texts from unseen languages, outperforming the best-performing unsupervised segmentation model \cite{glavavs2016unsupervised} by a wide margin.    



\section{CATS: Coherence-Aware Two-Level Transformer for Text Segmentation}

\begin{figure}
    \centering
    \includegraphics[scale=0.80]{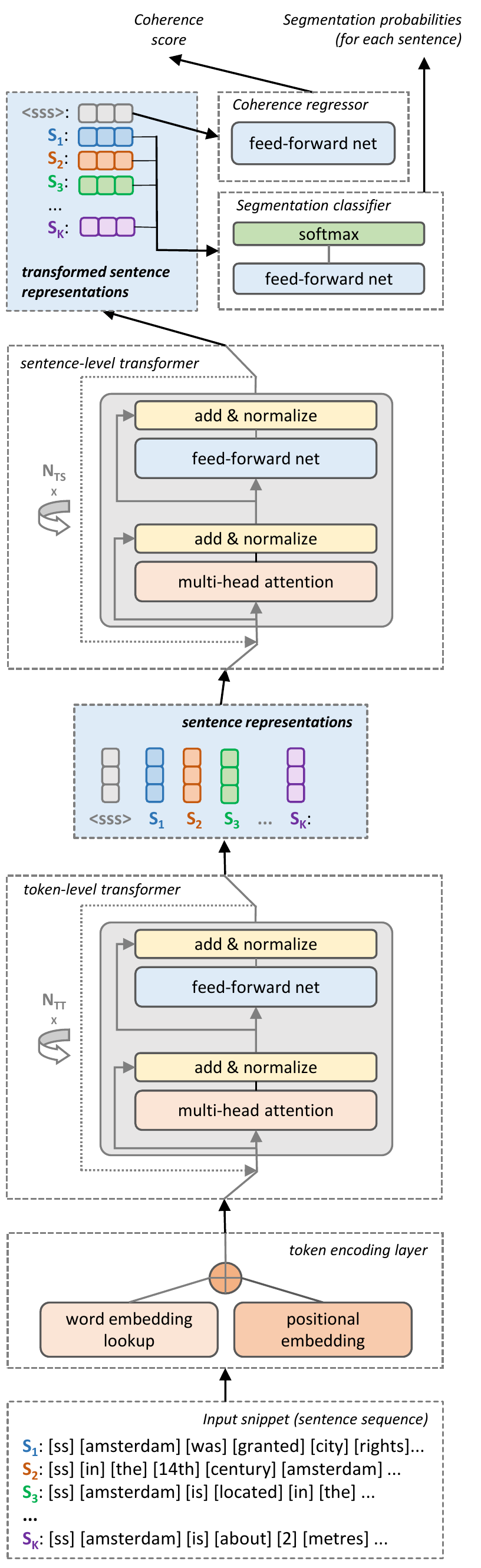}
    \caption{High-level depiction of the Coherence-Aware Text Segmentation (CATS) model.}
    \label{fig:model}
\end{figure}

Figure \ref{fig:model} illustrates the high-level architecture of the CATS model. 
A snippet of text -- a sequence of sentences of fixed length -- is an input to the model. Token encodings are a concatenation of a pretrained word embedding and a positional embedding. Sentences are first encoded from their tokens with a token-level Transformer \cite{vaswani2017attention}.
Next, we feed the sequence of obtained sentence representations 
to the second, sentence-level Transformer. Transformed (i.e., contextualized) sentence representations are next fed to the feed-forward segmentation classifier, which makes a binary segmentation prediction for each sentence. We additionally feed the encoding of the whole snippet (i.e., the sentence sequence) to the coherence regressor (a feed-forward net), which predicts a coherence score. In what follows, we describe each component in more detail.                
\subsection{Transformer-Based Segmentation}

The segmentation decision for a sentence clearly does not depend only on its content but also on its context, i.e., information from neighboring sentences. 
In this work, we employ the encoding stack of the attention-based Transformer architecture \cite{vaswani2017attention} to contextualize both token representations in a sentence and, more importantly, sentence representations within the snippet. We choose Transfomer encoders because (1) they have recently been reported to outperform recurrent encoders on a range of NLP tasks \cite{devlin2018bert,radford2018improving,shaw2018self} and (2) they are faster to train than recurrent nets.

\paragraph{Sentence Encoding.} Let $\mathbb{S} = \{S_1, S_2, \dots, S_K\}$ denote a single training instance -- a snippet consisting of $K$ sentences and let each sentence $S_i = \{t^i_1, t^i_2, \dots, t^i_T\}$ be a fixed-size sequence of $T$ tokens.\footnote{We trim/pad sentences longer/shorter than $T$ tokens.} Following \cite{devlin2018bert}, we prepend each sentence $S_i$ with a special sentence start token $t^i_0 = \text{[ss]}$, aiming to use the transformed representation of that token as the sentence encoding.\footnote{This eliminates the need for an additional self-attention layer for aggregating transformed token vectors into a sentence encoding.} We encode each token $t^i_j$ ($i \in \{1, \dots, K\}$, $j \in \{0, 1, \dots, T\}$) with a vector $\mathbf{t}^i_j$ which is the concatenation of a $d_e$-dimensional word embedding and a $d_p$-dimensional embedding of the position $j$. We use pretrained word embeddings and fix them in training; we learn positional embeddings as model's parameters. Let $\mathit{Transform}_T$ denote the encoder stack of the Transformer model \cite{vaswani2017attention}, consisting of $N_{TT}$ layers, each coupling a multi-head attention net with a feed-forward net.\footnote{For more details on the encoding stack of the Transformer model, see the original publication \cite{vaswani2017attention}.} We then apply $\mathit{Transform}_T$ to the token sequence of each snippet sentence: 
\begin{equation}
\{\mathbf{tt}^i_j\}^T_{j = 0} = \mathit{Transform}_{T}\left(\{\mathbf{t}^i_j\}^T_{j = 0}\right);
\end{equation}
\noindent The sentence encoding is then the transformed vector of the sentence start token [ss]: $\mathbf{s}_i = \mathbf{tt}^i_0$. 

\paragraph{Sentence Contextualization.} 

Sentence encodings $\{\mathbf{s}_i\}^K_{i = 1}$ produced with $\mathit{Transform}_T$ only capture the content of the sentence itself, but not its context. We thus employ a second, sentence-level Transformer $\mathit{Transform}_S$ (with $N_{TS}$ layers) to produce context-informed sentence representations. We prepend each sequence of non-contextualized sentence embeddings $\{\mathbf{s}_i\}^K_{i = 1}$ with a fixed embedding $\mathbf{s}_0$, denoting the snippet start token \textless sss\textgreater , in order to capture the encoding of the whole snippet (i.e., sequence of $K$ sentences) as the transformed embedding of the \textless sss\textgreater\,\,token: 
%
\begin{equation}
\{\mathbf{ss}_i\}^K_{i = 0} = \mathit{Transform}_{S}\left(\{\mathbf{s}_i\}^K_{i = 0}\right);
\end{equation}
\noindent with the transformed vector $\mathbf{ss}_0$ being the encoding of the whole snippet $\mathbb{S}$.

\paragraph{Segmentation Classification.} Finally, contextualized sentence vectors $\mathbf{ss}_i$ go into the segmentation classifier, a single-layer feed-forward net coupled with softmax function: 
\begin{equation}
\mathbf{\hat{y}}_i = \mathit{softmax}(\mathbf{ss}_i\mathbf{W}_{seg} + \mathbf{b}_{seg});   
\end{equation}
with $\mathbf{W}_{seg} \in \mathbb{R}^{(d_e + d_p) \times 2}$ and $\mathbf{b}_{seg} \in \mathbb{R}^2$ as classifier's parameters. Let $\mathbf{y}_i \in \{[0, 1], [1, 0]\}$ be the true segmentation label of the $i$-th sentence. The segmentation loss $J_{\mathit{seg}}$ is then the simple negative log-likelihood over all sentences of all $N$ snippets in the training batch:
\begin{equation}
    J_{\mathit{seg}} = -\sum^{N}_{n = 1}{\sum^{K}_{i = 1}{\ln \mathbf{\hat{y}}^n_i \cdot \mathbf{y}^n_i}}.
\end{equation}

\subsection{Auxiliary Coherence Modeling}

Given the obvious dependency between segmentation and coherence, we pair the segmentation task with an auxiliary task of predicting snippet coherence. 
To this effect, we couple each true snippet $\mathbb{S}$ from the original text with a corrupt (i.e., incoherent) snippet $\thickbar{\mathbb{S}}$, created by (1) randomly shuffling the order of sentences in $\mathbb{S}$ and (2) randomly replacing sentences from $\mathbb{S}$, with other document sentences. 

Let $(\mathbb{S},\thickbar{\mathbb{S}})$ be a pair of a true snippet and its corrupt counterpart, and ($\mathbf{ss}_0$, $\overline{\mathbf{ss}}_0$) their respective encodings, obtained with the Two-Level Transformer. The encodings of the correct snippet ($\mathbf{ss}_0$) and the scrambled snippet ($\overline{\mathbf{ss}_0}$) are then presented to the coherence regressor, which independently generates a coherence score for each of them. The scalar output of the coherence regressor is:
\begin{equation}
\mathbf{\hat{y}}_{\mathbb{S}} = \mathbf{ss}_0\mathbf{w}_{\mathit{c}} + b_{\mathit{c}};\hspace{1em} \mathbf{\hat{y}}_{\thickbar{\mathbb{S}}} = \overline{\mathbf{ss}_0}\mathbf{w}_{\mathit{c}} + b_{\mathit{c}};
\end{equation}
with $\mathbf{w}_{\mathit{c}} \in \mathbb{R}^{d_e + d_p}$ and $b_{\mathit{c}} \in \mathbb{R}$ as regressor's parameters. We then jointly softmax-normalize the scores for $\mathbb{S}$ and $\thickbar{\mathbb{S}}$:
%
\begin{equation}
    [\mathit{coh}(\mathbb{S}), \mathit{coh}(\thickbar{\mathbb{S}})] = \mathit{softmax}\left([\mathbf{\hat{y}}_{\mathbb{S}}, \mathbf{\hat{y}}_{\thickbar{\mathbb{S}}}]\right).
\end{equation}
We want to force the model to produce higher coherence score for the correct snippet $\mathbb{S}$ than for its corrupt counterpart $\thickbar{\mathbb{S}}$. We thus define the following contrastive margin-based coherence objective:
%
\begin{equation}
    J_\mathit{coh} = \max\hspace{-0.1em}\left(0, \delta_\mathit{coh} - (\mathit{coh}(\mathbb{S}) - \mathit{coh}(\hat{\mathbb{S}}))\right)
\end{equation}
where $\delta_\mathit{coh}$ is the margin by which we would like $\mathit{coh}(\mathbb{S})$ to be larger than $\mathit{coh}(\thickbar{\mathbb{S}})$.

\subsection{Creating Training Instances}

Our presumed training corpus contains documents that are generally longer than the snippet size $K$ and annotated for segmentation at the sentence level. 
We create training instances by sliding a sentence window of size $K$ over documents' sentences with a stride of $K/2$. 
%
For the sake of auxiliary coherence modeling, for each original snippet S, we create its corrupt
counterpart $\thickbar{\mathbb{S}}$ with the following corruption procedure: (1) we first randomly shuffle the order of sentences in $\mathbb{S}$; (2) for $p_1$ percent of snippets (random selection) we additionally replace sentences of the shuffled snippet (with the probability $p_2$) with randomly chosen sentences from other, non-overlapping document snippets.


\subsection{Inference}

At inference time, given a long document,  we need to make a binary segmentation decision for each sentence. Our model, however, does not take individual sentences as input, but rather sequences of $K$ sentences (i.e., snippets) and makes in-context segmentation prediction for each sentence. Since we can create multiple different sequences of $K$ consecutive sentences that contain some sentence $S$,\footnote{Sliding the sentence window with the stride of $1$, the $m$-th sentence will, in the general case, be found in $K$ different snippets: $[m-K+1:m]$, $[m-K+2:m+1]$, \dots, $[m:m+K-1]$.} our model can obtain multiple segmentation predictions for the same sentence. 
As we do not know apriori which of the snippets containing the sentence $S$ is the most reliable with respect to the segmentation prediction for $S$, we consider \textit{all} possible snippets containing $S$.  
In other words, at inference time, unlike in training, we create snippets by sliding the window of $K$ sentences over the document with the stride of $1$. 
Let $\mathcal{S} = \{\mathbb{S}_1, \mathbb{S}_2, \dots, \mathbb{S}_K\}$ be the set of (at most) $K$ different snippets containing a sentence $S$. We then average the segmentation probabilities predicted for the sentence $S$ over all snippets in $\mathcal{S}$:\footnote{The first element (i.e., index $[0]$) of the predicted vector $\hat{\mathbf{y}}$ denotes the (positive) segmentation probability.}
\begin{equation}
    P_{\mathit{seg}}(S) = \frac{1}{K}\sum_{\mathbb{S}_k \in \mathcal{S}}{\hat{\mathbf{y}}_S\left(\mathbb{S}_k\right)[0]}
\end{equation}
Finally, we predict that $S$ starts a new segment if $P_{\mathit{seg}}(S) > \tau$, where $\tau$ is the confidence threshold, tuned as a hyperparameter of the model. 

\label{sec:train_data}

\subsection{Cross-Lingual Zero-Shot Transfer}

Models that do not require any language-specific features other than pretrained word embeddings as input can (at least conceptually) be easily transferred to another language by means of a cross-lingual word embedding space \cite{Ruder2018survey,glavas2019properly}. 
Let $\mathbf{X}_{L1}$ be the monolingual embedding space of the source language (most often English), which we use in training and let $\mathbf{X}_{L2}$ be the independently trained embedding space of the target language to which we want to transfer the segmentation model. To transfer the model, we need to project target-language vectors from $\mathbf{X}_{L2}$ to the source-language space $\mathbf{X}_{L1}$. There is a plethora of recently proposed methods for inducing projection-based cross-lingual embeddings \cite[\textit{inter alia}]{faruqui2014improving,smith2017offline,artetxe2018robust,vulic2019really}. We opt for the supervised alignment model based on solving the Procrustes problem \cite{smith2017offline}, due to its simplicity and competitive performance in zero-shot language transfer of NLP models \cite{glavas2019properly}. 
Given a limited-size word translation training dictionary $D$, we obtain the linear projection matrix $\mathbf{W}_{L2\rightarrow L1}$ between $\mathbf{X}_{L2}$ and $\mathbf{X}_{L1}$ as follows: 
\begin{equation}
    \mathbf{W}_{L2\rightarrow L1} = \mathbf{UV}^\top;\,
    \hspace{0.2em} \mathbf{U\Sigma V}^\top = \mathit{SVD}(\mathbf{X}_{S} {\mathbf{X}_{T}}^\top);
    \label{eq:proc}
\end{equation}
with $\mathbf{X}_{S} \subset \mathbf{X}_{L1}$ and $\mathbf{X}_{T} \subset \mathbf{X}_{L2}$ as subsets of monolingual spaces that align vectors from training translations pairs from $D$. Once we obtain $\mathbf{W}_{L2\rightarrow L1}$, the language transfer of the segmentation model is straightforward: we input the embeddings of L2 words from the projected space $\mathbf{X}'_{L2} = \mathbf{X}_{L2} \mathbf{W}_{L2\rightarrow L1}$.

\section{Experimental Setup}

We first describe datasets used for training and evaluation and then provide the details on the comparative evaluation setup and model optimization.

\subsection{Data}

\paragraph{\textsc{Wiki-727K} Corpus.} \newcite{koshorek2018text} leveraged the manual structuring of Wikipedia pages into sections to automatically create a large segmentation-annotated corpus. 
\textsc{Wiki-727K} consists of 727,746 documents created from English (EN) Wikipedia pages, divided into training (80\%), development (10\%), and test portions (10\%). We train, optimize, and evaluate our models on respective portions of the \textsc{Wiki-727K} dataset. 


\paragraph{Standard Test Corpora.} \newcite{koshorek2018text} additionally created a small evaluation set \textsc{Wiki-50} to allow for comparative evaluation against unsupervised segmentation models, e.g., the \textsc{GraphSeg} model of \newcite{glavavs2016unsupervised}, for which evaluation on large datasets is prohibitively slow. For years, the synthetic dataset of \newcite{choi2000advances} was used as a standard becnhmark for text segmentation models. \textsc{Choi} dataset contains 920 documents, each of which is a concatenation of 10 paragraphs randomly sampled from the Brown corpus. \textsc{Choi} dataset is divided into subsets containing only documents with specific variability of segment lengths (e.g., segments with 3-5 or with 9-11 sentences).\footnote{Following \newcite{koshorek2018text}, we evaluate our models on the whole \textsc{Choi} corpus and not on specific subsets.} Finally, we evaluate the performance of our models on two small datasets, \textsc{Cities} and \textsc{Elements}, created by \newcite{chen2009global} from Wikipedia pages dedicated to the cities of the world and chemical elements, respectively.

\paragraph{Other Languages.} In order to test the performance of our Transformer-based models in zero-shot language transfer setup, we prepared small evaluation datasets in other languages. Analogous to the \textsc{Wiki-50} dataset created by \newcite{koshorek2018text} from English (EN) Wikipedia, we created \textsc{Wiki-50-CS}, \textsc{Wiki-50-FI}, and \textsc{Wiki-50-TR} datasets consisting of 50 randomly selected pages from Czech (CS), Finnish (FI), and Turkish (TR) Wikipedia, respectively.\footnote{For our language transfer experiments we selected target languages from different families and linguistic typologies w.r.t English as our source language: Czech is, like English, an Indo-European language (but as a Slavic language it is, unlike English, fusional by type); Finnish is an Uralic language (fusionally-agglutinative by type); whereas Turkish is a Turkic language (agglutinative by type).}    

\subsection{Comparative Evaluation}

\paragraph{Evaluation Metric.} Following previous work \cite{riedl2012topictiling,glavavs2016unsupervised,koshorek2018text}, we also adopt the standard text segmentation measure $P_k$ \cite{beeferman1999statistical} as our evaluation metric. $P_k$ score is the probability that a model makes a wrong prediction as to whether the first and last sentence of a randomly sampled snippet of $k$ sentences belong to the same segment (i.e., the probability of the model predicting the same segment for the sentences from different segment or different segments for the sentences from the same segment). Following \cite{glavavs2016unsupervised,koshorek2018text}, we set $k$ to the half of the average ground truth segment size of the dataset.

\paragraph{Baseline Models.} We compare CATS against the state-of-the-art  neural segmentation model of \citet{koshorek2018text} and against \textsc{GraphSeg} \cite{glavavs2016unsupervised}, 
 the state-of-the-art unsupervised text segmentation model. Additionally, as a sanity check, we evaluate the \textsc{Random} baseline -- it assigns a positive segmentation label to a sentence with the probability that corresponds to the ratio of the total number of segments (according to the gold segmentation) and total number of sentences in the dataset.        

\subsection{Model Configuration}

\paragraph{Model Variants.} We evaluate two variants of our two-level transformer text segmentation model: with and without the auxiliary coherence modeling. The first model, TLT-TS, minimizes only the segmentation objective $J_\mathit{seg}$. CATS, our second model, is a multi-task learning model that alternately minimizes the segmentation objective $J_\mathit{seg}$ and the coherence objective $J_\mathit{coh}$. We adopt a balanced alternate training regime for CATS in which a single parameter update based on the minimization of  $J_\mathit{seg}$ is followed by a single parameter update based on the optimization of $J_\mathit{coh}$. 

\paragraph{Word Embeddings.} In all our experiments we use 300-dimensional monolingual \textsc{fastText} word embeddings pretrained on the Common Crawl corpora of respective languages: EN, CS, FI, and TR.\footnote{\url{https://tinyurl.com/y6j4gh9a}} We induce a cross-lingual word embedding space, needed for the zero-shot language transfer experiments, by projecting CS, FI, and TR monolingual embedding spaces to the EN embedding space. Following \cite{smith2017offline,glavas2019properly}, we create training dictionaries $D$ for learning projection matrices by machine translating 5,000 most frequent EN words to CS, FI, and TR. 

\paragraph{Model Optimization.} 

We optimize all hyperparameters, including the data preparation parameters like the snippet size $K$, via cross-validation on the development portion of the Wiki-727K dataset. We found the following configuration to lead to robust\footnote{Given the large hyperparameter space and large training set, we only searched over a limited-size grid of hyperparameter configurations. It is thus likely that a better-performing configuration than the one reported can be found with a more extensive grid search.} performance for both TLT-TS and CATS: (1) training instance preparation: snippet size of $K = 16$ sentences with $T = 50$ tokens; scrambling probabilities $p_1 = p_2 = 0.5$; (2) configuration of Transformers: $N_{TT} = N_{TS} = 6$ layers and with $4$ attention heads per layer in both transformers;\footnote{We do not tune other transformer hyperparameters, but rather adopt the recommended values from \cite{vaswani2017attention}: filter size of $1024$ and dropout probabilities of $0.1$ for both attention layers and feed-forward ReLu layers.} (3) other model hyperparameters: positional embedding size of $d_p = 10$; coherence objective contrastive margin of $\delta_\mathit{coh} = 1$. We found different optimal inference thresholds: $\tau = 0.5$ for the segmentation-only TLT-TS model and $\tau = 0.3$ for the coherence-aware CATS model. We trained both TLT-TS and CATS in batches of $N = 32$ snippets (each with $K = 16$ sentences), using the Adam optimization algorithm \cite{kingma2014adam} with the initial learning rate set to $10^{-4}$.

\setlength{\tabcolsep}{12pt}
\begin{table*}[t]
\centering
{
\begin{tabularx}{\linewidth}{l l c c c c c}
\toprule 
Model & Model~Type & \textsc{Wiki-727K} & \textsc{Wiki-50} & \textsc{Choi} & \textsc{Cities} & \textsc{Elements} \\ \midrule
\textsc{Random} & unsupervised & 53.09 & 52.65 & 49.43 & 47.14 & 50.08 \\
\textsc{GraphSeg} & unsupervised & -- & 63.56 & \textbf{5.6}--\textbf{7.2}* & 39.95 & 49.12 \\
\newcite{koshorek2018text} & supervised & 22.13 & 18.24 & 26.26 & 19.68 & 41.63 \\ \midrule
TLT-TS & supervised & 19.41 & 17.47 & 23.26 & 19.21 & 20.33 \\
CATS & supervised & \textbf{15.95} & \textbf{16.53} & 18.50 & \textbf{16.85} & \textbf{18.41} \\
\bottomrule
\end{tabularx}
}
\caption{Performance of text segmentation models on five English evaluation datasets. 
\textsc{GraphSeg} model \cite{glavavs2016unsupervised} was evaluated independently on different subcorpora of the \textsc{Choi} dataset (indicated with an asterisk).}
\label{tbl:base}
\end{table*}

\section{Results and Discussion}

We first present and discuss the results that our models, TLT-TS and CATS, yield on the previously introduced EN evaluation datasets. We then report and analyze models' performance in the cross-lingual zero-shot transfer experiments. 

\subsection{Base Evaluation}

Table \ref{tbl:base} shows models' performance on five EN evaluation datasets. 
Both our Transformer-based models -- TLT-TS and CATS -- outperform the competing supervised model of \citet{koshorek2018text}, a hierarchical encoder based on recurrent components, across the board. The improved performance that TLT-TS has with respect to the model of \newcite{koshorek2018text} is consistent with improvements that Transformer-based architectures yield in comparison with models based on recurrent components in other NLP tasks \cite{vaswani2017attention,devlin2018bert}. The gap in performance is particularly wide ($>$20 $P_k$ points) for the \textsc{Elements} dataset. Evaluation on the \textsc{Elements} test set is, arguably, closest to a true domain-transfer setting:\footnote{The \textsc{Choi} dataset -- albeit from a different domain -- is synthetic, which impedes direct performance comparisons with other evaluation datasets.} while the train portion of the \textsc{Wiki-727K} set contains pages similar in type to those found in \textsc{Wiki-50} and \textsc{Cities} test sets, 
it does not contain any Wikipedia pages about chemical elements (all such pages are in the \textsc{Elements} test set). This would suggest that TLT-TS and CATS offer more robust domain transfer than the recurrent model of \newcite{koshorek2018text}.         

CATS significantly\footnote{According to the non-parametric random shuffling test \cite{yeh2000more}: $p < 0.01$ for \textsc{Wiki-727K}, \textsc{Choi} and \textsc{Cities}; $p < 0.05$ for \textsc{Wiki-50} and \textsc{Elements}.} and consistently outperforms TLT-TS. This empirically confirms the usefulness of explicit coherence modeling for text segmentation. Moreover, \newcite{koshorek2018text} report human performance on the \textsc{Wiki-50} dataset of $14.97$, which is a mere one $P_k$ point better than the performance of our coherence-aware CATS model.  

The unsupervised \textsc{GraphSeg} model of \newcite{glavavs2016unsupervised} seems to outperform all supervised models on the synthetic \textsc{Choi} dataset. We believe that this is primarily because (1) by being synthetic, the \textsc{Choi} dataset can be accurately segmented based on simple lexical overlaps and word embedding similarities (and \textsc{GraphSeg} relies on similarities between averaged word embeddings) and because (2) by being trained on a much more challenging real-world \textsc{Wiki-727K} dataset -- on which lexical overlap is insufficient for accurate segmentation -- supervised models learn to segment based on deeper natural language understanding (and learn not to encode lexical overlap as reliable segmentation signal). Additionally, \textsc{GraphSeg} is evaluated separately on each subset of the \textsc{Choi} dataset, for each of which it is provided the (gold) minimal segment size, which further facilitates and improves its predicted segmentations.          

\subsection{Zero-Shot Cross-Lingual Transfer}

In Table \ref{tbl:cl} we show the results of our zero-shot cross-lingual transfer experiments. In this setting, we use our Transformer-based models, trained on the English \textsc{Wiki-727K} dataset, to segment texts from the \textsc{Wiki-50-X} (X $\in$ \{CS, FI, TR\}) datasets in other languages. 
As a baseline, we additionally evaluate \textsc{GraphSeg} \cite{glavavs2016unsupervised}, as a language-agnostic model requiring only pretrained word embeddings of the test language as input.     

\setlength{\tabcolsep}{14pt}
\begin{table}[t]
\centering
{
\begin{tabularx}{\linewidth}{l c c c}
\toprule 
\textit{Model} & CS & FI & TR \\ \midrule
\textsc{Random} & 52.92 & 52.02 & 45.04 \\
\textsc{GraphSeg} & 49.47 & 49.28 & 39.21 \\ \midrule
TLT-TS & 24.27 & 25.99 & 25.89 \\
CATS & \textbf{22.32} & \textbf{22.87} & \textbf{24.20} \\
\bottomrule
\end{tabularx}
}
\caption{Performance of text segmentation models in zero-shot language transfer setting on the \textsc{Wiki-50-X} (X $\in$ \{CS, FI, TR\}) datasets.}
\label{tbl:cl}
\end{table}

Both our Transformer-based models, TLT-TS and CATS, outperform the unsupervised \textsc{GraphSeg} model (which seems to be only marginally better than the random baseline) by a wide margin. The coherence-aware CATS model is again significantly better ($p < 0.01$ for FI and $p < 0.05$ for CS and TR) than the TLT-TS model which was trained to optimize only the segmentation objective. While the results on the \textsc{Wiki-50}-\{CS, FI, TR\} datasets are not directly comparable to the results reported on the EN \textsc{Wiki-50} (see Table \ref{tbl:base}) because the datasets in different languages do not contain mutually comparable Wikipedia pages, results in Table \ref{tbl:cl} still suggest that the drop in performance due to the cross-lingual transfer is not big. This is quite encouraging as it suggests that it is possible to, via the zero-shot language transfer, rather reliably segment texts from under-resourced languages lacking sufficiently large gold-segmented data needed to directly train language-specific segmentation models (that is, robust neural segmentation models in particular).
\section{Related Work}

In this work we address the task of text segmentation -- we thus provide a detailed account of existing segmentation models. Because our CATS model has an auxiliary coherence-based objective, we additionally provide a brief overview of research on modeling text coherence.   

\subsection{Text Segmentation}

Text segmentation tasks come in two main flavors: (1) linear (i.e., sequential) text segmentation and (2) hierarchical segmentation in which top-level segments are further broken down into sub-segments. While the hierarchical segmentation received a non-negligible research attention \cite{yaari1997,eisenstein2009,du2013topic}, the vast majority of the proposed models (including this work) focus on linear segmentation \cite[\textit{inter alia}]{hearst1994multi,beeferman1999statistical,choi2000advances,brants2002topic,misra2009,riedl2012topictiling,glavavs2016unsupervised,koshorek2018text}. 


In one of the pioneering segmentation efforts, \newcite{hearst1994multi} proposed an unsupervised TextTiling algorithm based on the lexical overlap between adjacent sentences and paragraphs. \newcite{choi2000advances} computes the similarities between sentences in a similar fashion, but renormalizes them within the local context; the segments are then obtained through divisive clustering. \newcite{utiyama2001statistical} and \newcite{fragkou2004} minimize the segmentation cost via exhaustive search with dynamic programming. 

Following the assumption that topical cohesion guides the segmentation of the text, a number of segmentation approaches based on topic models have been proposed. \newcite{brants2002topic} induce latent representations of text snippets using probabilistic latent semantic analysis \cite{hofmann1999probabilistic} and segment based on similarities between latent representations of adjacent snippets. \newcite{misra2009} and \newcite{riedl2012topictiling} leverage topic vectors of snippets obtained with the Latent Dirichlet Allocation model \cite{blei2003latent}. While \newcite{misra2009} finds a globally optimal segmentation based on the similarities of snippets' topic vectors using dynamic programming, \newcite{riedl2012topictiling} adjust the TextTiling model of \cite{hearst1994multi} to use topic vectors instead of sparse lexicalized representations of snippets.

\newcite{malioutov2006minimum} proposed a first graph-based model for text segmentation. They segment lecture transcripts by first inducing a fully connected sentence graph with edge weights corresponding to cosine similarities between sparse bag-of-word sentence vectors and then running a minimum normalized multiway cut algorithm to obtain the segments. \newcite{glavavs2016unsupervised} propose \textsc{GraphSeg}, a graph-based segmentation algorithm similar in nature to \cite{malioutov2006minimum}, which uses dense sentence vectors, obtained by aggregating word embeddings, to compute intra-sentence similarities and performs segmentation based on the cliques of the similarity graph. 

Finally, \newcite{koshorek2018text} identify Wikipedia as a free large-scale source of manually segmented texts that can be used to train a supervised segmentation model. They train a neural model that hierarchically combines two bidirectional LSTM networks and report massive improvements over unsupervised segmentation on a range of evaluation datasets. The model we presented in this work has a similar hierarchical architecture, but uses Transfomer networks instead of recurrent encoders. Crucially, CATS additionally defines an auxiliary coherence objective, which is coupled with the (primary) segmentation objective in a multi-task learning model.

\subsection{Text Coherence}

Measuring text coherence amounts to predicting a score that indicates how meaningful the order of the information in the text is.
The majority of the proposed text coherence models are grounded in formal theories of text coherence, among which the entity grid model \cite{barzilay2008modeling}, based on the centering theory of \newcite{grosz1995centering}, is arguably the most popular. The entity grid model represent texts as matrices encoding the grammatical roles that the same entities have in different sentences. The entity grid model, as well as its extensions \cite{elsner2011extending,feng2012extending,feng2014impact,nguyen2017neural} require text to be preprocessed -- entities extracted and grammatical roles assigned to them -- which prohibits an end-to-end model training. 

In contrast, \newcite{li2014model} train a neural model that couples recurrent and recursive sentence encoders with a convolutional encoder of sentence sequences in an end-to-end fashion on limited-size datasets with gold coherence scores. Our models' architecture is conceptually similar, but we use Transformer networks to both encode sentences and sentence sequences. With the goal of supporting text segmentation and not aiming to predict exact coherence scores, our model does not require gold coherence labels; instead we devise a coherence objective that contrasts original text snippets against corrupted sentence sequences.          
\section{Conclusion}

Though the segmentation of text depends on its (local) coherence, existing segmentation models capture coherence only implicitly via lexical or semantic overlap of (adjacent) sentences. 
In this work, we presented CATS, a novel supervised model for text segmentation that couples segmentation prediction with explicit auxiliary coherence modeling. CATS is a neural architecture consisting of two hierarchically connected Transformer networks: the lower-level sentence encoder generates input for the higher-level encoder of sentence sequences. We train the model in a multi-task learning setup by learning to predict (1) sentence segmentation labels and (2) that original text snippets are more coherent than corrupt sentence sequences. We show that CATS yields state-of-the-art performance on several text segmentation benchmarks and that it can -- in a zero-shot language transfer setting, coupled with a cross-lingual word embedding space -- successfully segment texts from target languages unseen in training. 

Although effective for text segmentation, our coherence modeling is still rather simple: we use only fully randomly shuffled sequences as examples of (highly) incoherent text. In subsequent work, we will investigate negative instances of different degree of incoherence as well as more elaborate objectives for (auxiliary) modeling of text coherence.

\bibliography{references.bib}
\bibliographystyle{aaai}

\end{document}